\documentclass[nohyperref]{article}

\usepackage{microtype}
\usepackage{graphicx}
\usepackage{booktabs} 

\usepackage{hyperref}

\usepackage[accepted]{icml2023}

\usepackage{amsmath}
\usepackage{amssymb}
\usepackage{mathtools}
\usepackage{amsthm}
\usepackage{bm}
\usepackage{subfig}

\usepackage[capitalize,noabbrev]{cleveref}

\theoremstyle{plain}

\theoremstyle{definition}

\theoremstyle{remark}

\usepackage[textsize=tiny]{todonotes}

\icmltitlerunning{Shapley Based Residual Decomposition for Instance Analysis}

\begin{document}

\twocolumn[
\icmltitle{Shapley Based Residual Decomposition for Instance Analysis}



\icmlsetsymbol{equal}{*}

\begin{icmlauthorlist}
\icmlauthor{Tommy Liu}{anu}
\icmlauthor{Amanda Barnard}{anu}
\end{icmlauthorlist}

\icmlaffiliation{anu}{School of Computing, Australian National University, Canberra, Australia}

\icmlcorrespondingauthor{Tommy Liu}{tommy.liu@anu.edu.au}

\icmlkeywords{Machine Learning, ICML}

\vskip 0.3in
]



\printAffiliationsAndNotice{} 

\begin{abstract}
In this paper, we introduce the idea of decomposing the residuals of regression with respect to the data instances instead of features. This allows us to determine the effects of each individual instance on the model and each other, and in doing so makes for a model-agnostic method of identifying instances of interest. In doing so, we can also determine the appropriateness of the model and data in the wider context of a given study. The paper focuses on the possible applications that such a framework brings to the relatively unexplored field of instance analysis in the context of Explainable AI tasks. 
\end{abstract}

\section{Introduction and Related work}

Often, the task of identifying instances of interest from particular data lies with individual researchers' intuitions, experience, or knowledge. It is typically the case that we place importance on some kinds of data more than others. It is therefore desirable to quantify the impact that each instance has on the model, and each other for a given task. The tasks of determining outliers and influential instances have been extensively studied in Statistics \cite{cook1979influential, hawkins1980identification, hadi2009detection, sullivan2021so}, Databases Analysis \cite{angiulli2009detecting}, Machine Learning \cite{escalante2005comparison, breunig2000lof, liu2008isolation}, and more. This issue however remains context dependent and difficult to approach for general cases, especially in many contemporary machine learning problems where explainability is highly desired \cite{shin2021effects, doshi2017towards}. 
\\
\\
Explainable design of algorithms and techniques in the field of Machine Learning has seen significant research interest in recent years \cite{doshi2017towards, miller2019explanation,das2020opportunities}. In particular, post-hoc explanations such as LIME \cite{ribeiro2016should} and SHAP \cite{lundberg2017unified} have seen explosive growth from both academic and commercial interests. There now exist many such methods for post-hoc explanations which suit a variety of different needs, particularly over the feature space. However, it is the case that explanations and analysis tools over the instance space remain relatively unexplored. Traditionally the task of instance explanations has been neglected in favour of the computationally simpler task of understanding how the features combine instead.
\\
\\
There still remain no clear methods to understand how instances contribute to a model, particularly in the presence of possibly degenerate data. The needs techniques to analyse data-sets and the instance space remain to spread out over many sub-fields such as Outlier and Anomaly detection, Instance Selection, and Statistical Influence Measures among many others \cite{escalante2005comparison, cook1982residuals, pope1976statistics, li2017feature}. Many of the algorithms in these fields are designed for specific tasks meaning that it remains difficult for a data analyst to determine where to start probing the data. The task of analysing such instances has many challenges that must be overcome, in particular this lack of unifying definition across the various sub-fields mentioned in the context of Machine Learning tasks. A particularly relevant method that provides insights into instances using Shapley Values is that of Data Shapley values \cite{ghorbani2019data, jia2019towards} and Distributional Data Shapley values \cite{ghorbani2020distributional}, which decompose the contribution of each instance to the loss term of a model to determine a final `value' for each instance. This `value' of an instance can then be used to determine how valuable or useful an instance or similar instances may be (for the purposes of data valuation tasks such as data warehousing). However, the `value' of an instance is derived from three sources: the algorithm; data; and the choice of valuation method. Little insight is provided into how the interaction of these three plays into the final `value' of this data.
\\
\\
In this paper, we aim to demonstrate how choices regarding the analysis of instances can be aided with contemporary explainability frameworks such as the Shapley Value \cite{lundberg2017unified, shapley1953value}. We introduce a method to generally analyse data instances based on this Shapley Value and focuses on instance behaviours instead of properties. This is done by choosing the value function such that it provides insights into both the data and the model. Our method provides evidence for instances or patterns across the data-set to be considered interesting for a given model. The term interesting may include categories such as outliers, anomalies, influential instances, or exception data in general. The actual cause of these patterns will typically be data-set and domain-specific, our method provides a starting point to determine which instances need further analysis, along with how instances combine to produce final outputs. This interpretation allows for a more direct approach to screening methods instead of having multiple data analysis stages. One aspect of this work could be considered an extension to Data Shapley values in that it not only identifies data instances with large effects but also provides more insights into how these effects are distributed across the data and the relationships present within the data itself. 
\\
\\
In this paper, our contributions are as follows:
\begin{itemize}
    \item We introduce the \textit{Residual Decomposition} framework for the detection of interesting instances and their behaviours in the context of a given model. We also make use of known approaches to speed up the computation of these values.
    \item We present several newly derived quantities and methods from our technique which can provide insights into the behaviour and nature of data instances and their interactions.
    \item We demonstrate with examples of testing data how previously unknown samples can be selected and how or why they may be interesting by means of the novel CC plot, along with the contribution and composition quantities.
\end{itemize}

Previously, Shapley Values have seen significant usage in many areas of data analysis, some examples include: explaining how features combine to produce the final outputs \cite{lundberg2017unified, huang2023explainable}, determining the importance of particular features \cite{cohen2005feature, liu2021fast, fryer2021shapley}, determining how much value each instance has in data valuation \cite{ghorbani2019data, jia2019towards}, explaining how or why anomalies arise \cite{takeishi2020anomaly}, and more. One reason that Shapley Values have seen such popularity is because they satisfy several elegant axiomatic properties such as Fairness, Additivity, and Rationality.

\subsection{Shapley Values and Features}

Shapley Values \cite{shapley1953value} are a game theoretical method of dividing the total value of a group amongst its individual members. Formally speaking, over a set $F$ of individuals, the Shapley value $\phi_i$ of the $i^{th}$ individual for a given method of evaluation $v(\cdot)$ is given by:
\begin{equation}
    \phi_i = \sum_{S\subset F\backslash \{i\} } \frac{|S|!(|F| - |S| - 1)!}{|F|!} [v(S \cup  \{i\})  - v(S)]
\end{equation}
Where the sum is evaluated over possible subsets $S \subset F \backslash \{i\}$. Over recent years, Shapley values have gained popularity in Machine Learning communities as a method of post-hoc model analysis into how features combine to produce outputs. For example, by setting $F$ as the set of features in a given data-set and $v$ as the evaluation of outputs. One choice of $v$ is the conditional expectation of a model $f$ given some known values for features. The most popular framework using Shapley Values for feature explanations is the SHapley Additive exPlanations (SHAP) \cite{lundberg2017unified} which provides both model-specific and agnostic approaches and provides insights into how features combine to produce the final model outputs.

\subsection{Shapley Values for Instances}

By choosing different sets and valuations, Shapley Values can be applied over data instances. The difference in interpretation is simply how instances combine to produce some metric over model outputs. Shapley Values have been applied for Data Valuation contexts, by setting the values for $F$ to be the instances in the data-set. This is done with a valuation function such as the loss (i.e., mean squared error) over some evaluation set (the test set) with respect to a given model \cite{ghorbani2019data}.  This application of Shapley Values over the instance space has largely been focused on the task of determining the value of particular data instances, i.e., for the purposes of data exchange and warehousing \cite{ghorbani2019data}.
\\
\\ 
Another approach that makes use of Shapley Values over data instances was introduced in the Extended Shapley framework, this framework attributes responsibility for the loss to both the data and model \cite{yona2021s}. This approach similarly makes use of an evaluation function in decomposing the contribution of not only the instances but also the model towards the model performance (i.e., loss term) but provides little insight into the interactions between instances. These existing approaches only identify a single term representing the value of an individual instance. These approaches provide little insight into how or why such instances may be important since they are mainly focused on the loss term of a particular model evaluation. It is therefore desirable to extend the valuation function to provide precisely these insights into the interactions within the data.

\section{Methodology}

\begin{figure}[!t]
    \centering
    \includegraphics[width=\textwidth/2]{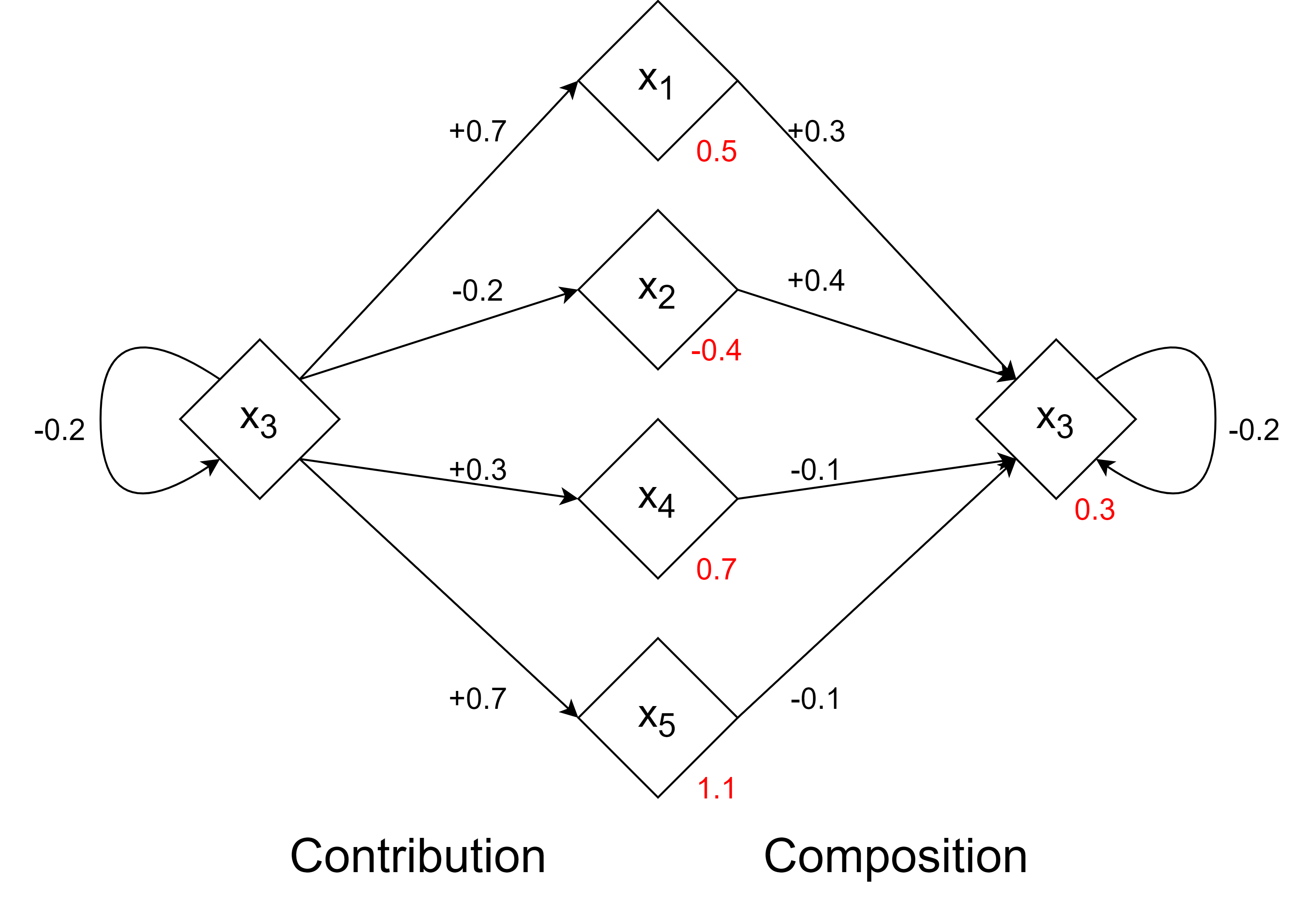}
    \caption{Figure showing example of the contribution of instance $x_3$ to other instances, and composition from other instances to $x_3$. Red values indicate the residual value for a particular instance $x_i$.}
    \label{fig:ccexample}
\end{figure}
\begin{figure}[!h]
    \centering
    \includegraphics[width=\textwidth/2]{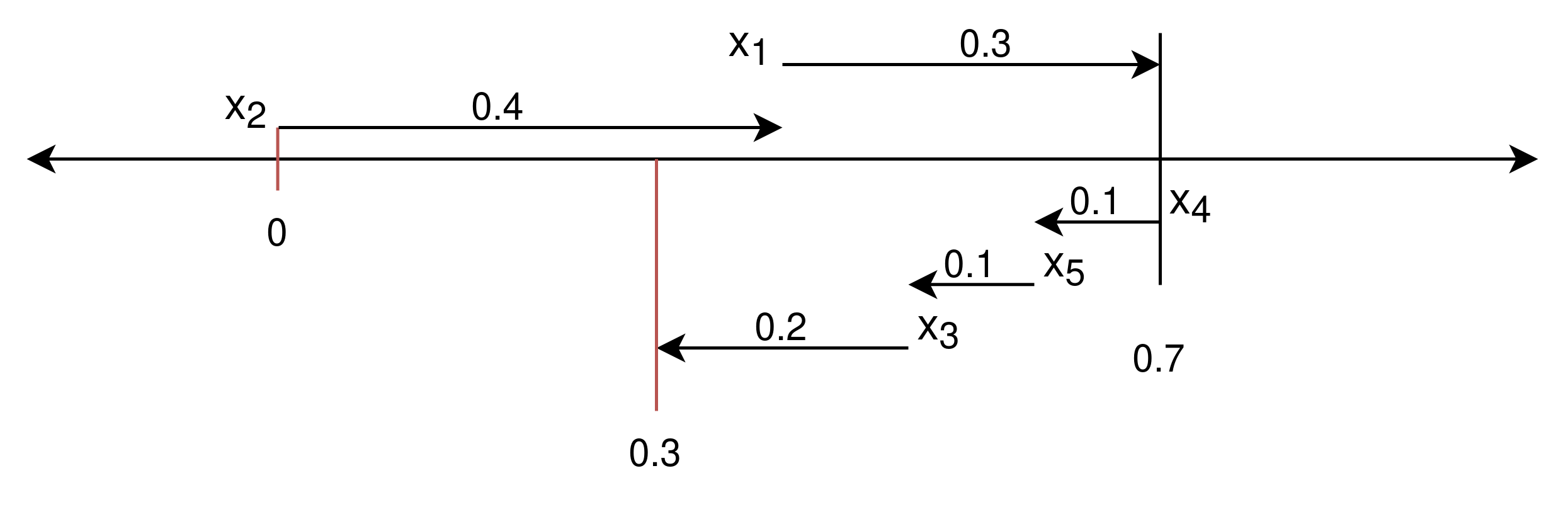}
    \caption{Figure showing total residual of instance $x_3$ with final residual of 0.3, with total positive effects from $x_1, x_2$ and negative effects from $x_3,x_4,x_5$.}
    \label{fig:iexample}
\end{figure}
Let $X$ be a data-set of $n$ individuals $X = \{x_1,...,x_n\}$ and $Y = \{y_1,...,y_n\}$ a set of corresponding labels. A typical real-value loss term for a machine learning method such as Mean Squared Error ($\frac{1}{n}\sum_{i=1}^n e_i^2$) is composed of $n$ individual residual terms $e_i = f(x_i) - y_i$ for some function $f$. In the same vein, we seek to decompose the residuals $e_i$ further into $n$ components $\phi_{i,j}$ such that $\sum_{j}^n \phi_{i,j} = e_i$. This property can be exactly satisfied by the Shapley Values of the residuals $e_i$. Well-established algorithms for computing the Shapley Value value can be modified to produce the desired result by changing the functional form of the value function. Our benchmark approach is that of the (truncated) permutation sampling based Monte Carlo algorithm \cite{ghorbani2019data, castro2009polynomial}. This approximation formula is provided in Equation \ref{eq:MCShapley}. 
\begin{equation}\label{eq:MCShapley}
\bm{\phi_i} = \mathbb{E}_{\pi \sim \Pi}[v(S_\pi^i\cup \{i\}) - v(S_\pi^i)]
\end{equation}
Where $\pi$ is a permutation of a uniform distribution over all $n!$ permutations of the data points $\Pi$, and $S_\pi^i$ is the set of instances before datum $i$ in permutation $\pi$. The algorithm involves scanning over a permutation in each iteration to get the value of the $i^{th}$ instance and taking the sample mean over some fixed number of permutation evaluations. This approach has been demonstrated to see convergence on the order $n$ (typically $3n$) \cite{ghorbani2019data, maleki2013bounding} and some chosen cut-off can be selected to stop the algorithm when the value change is smaller than the cut-off value. We choose the value function $v$ in this approximation to be the residual values over some fixed data-set $X \in \mathbb{R}^{n\times p}$ for a model fitted using only those data instances $S_\pi^i$ given by Equation  \ref{eq:residualvalue}. 
\begin{equation}\label{eq:residualvalue}
v(S_\pi^i) = \{f_{S_\pi^i} (x_j) - y_j\}_{j=1}^n
\end{equation}
Where $f_{S_\pi^i}(\cdot)$ is the model trained on only those instances in $S_\pi^i$. The sets $S$ and $X$ may be disjoint if we wish to evaluate what we term the asymmetric case, and symmetric case if $S=X$. We obtain a set of Shapley Values put into matrix form $\bm{\phi} \in \mathbb{R}^{|X|\times |S|}$, which recover the residual contribution of an instance $i$ to all other instances $j$. That is, by the efficiency property of Shapley Values we will have $\sum_{j=1}^n \phi_{i,j} = e_i$. We dub this the Shapley Values that a residual is \textit{composed} of. By contrast, the Shapley Values $\phi_{j,0},...,\phi_{j,N}$ are how much a particular instance $j$ \textit{contributed} to the residuals of all other data instances. Note that the composition sum or mean is a direct analogy to the residual values (by the additivity principle). In the case that our training set is equal to our testing set and $i=j$, the contribution-composition matrices are square (symmetric case) and this framework aims to describe the two-way relationships between instances, along with patterns in the data. When $i\neq j$ (asymmetric case) then we are more interested in the relationships between the instances present within the training and testing sets (i.e. for the task of data valuation or instance selection).
\\
\\
The permutation sampling method however scales poorly as the total time consumed significantly increases with the number of instances in the data, and the complexity of the model itself may be significantly higher than order $n$. Lundberg et al., introduced the KernelSHAP algorithm which approximates the Shapley Values by means of a weighted least squares regression taking the form of equation \ref{eq:wls} \cite{lundberg2017unified}.
\begin{equation}\label{eq:wls}
	\sum_{S\subseteq M} \left( v(S) - (\phi_0 + \sum_{j\in S}\phi_j) \right)^2 \cdot k(M, S)
\end{equation}
This describes the `optimal' use of the sampling budget according to the Shapley Kernel $K(M, S)$, meaning that it is possible to get the `best' approximation to the Shapley Value given a certain computational budgetary constraint. Under our choice of the value function, the base value $phi_0$ is eliminated. This is because under KernelSHAP the base value represents the ``expected value predicted by the model if no features were known", the expected value under the residuals interpretation is always 0 since across a number of subsets, the expected value of an unknown data point is exactly the model predicted value. 
\\
\\
For a large number of settings where the number of data points is large, or model training is highly expensive, the convergence on the order $n * $model complexity is still relatively infeasible. The Gradient-Shapley (G-Shapley) was previously proposed to speed up the computation of Data Shapley values which works for variations of models making use of Stochastic Gradient Descent (SGD). G-Shapley considers the values for a single epoch by updating the instances in a permutation by performing SGD on these single instances. Another similar strategy to approximate the Shapley values, particularly in SGD settings is that of influence functions explored by \cite{jia2019towards}. Influence functions have seen significant interest in Machine Learning and Explainable AI recently \cite{koh2017understanding} and can greatly reduce the amount of model retraining required by describing the change in the model when an instance is removed (the leave-one-out error) without the computational process of retraining the model. We explore the computational and approximation costs further in section \ref{sec.runtime.performance}.

\subsection{Interpretation of Contribution and Composition}

An example of an individual instance's contribution and composition values is illustrated in Figure \ref{fig:ccexample} for a hypothetical example data-set. A contribution towards a residual that makes its magnitude smaller is considered a positive effect and negative otherwise. For example, the relationship $x_3 \rightarrow x_1$ has a negative, since it decreases the magnitude of the relationship, whereas $x_3 \rightarrow x_4$ is a positive since it increases the value of the residual. We therefore recalculate the contribution values which we denote $\phi^c_{i,j}$ using the normalization given in Equation \ref{eq:econtributionnorm}. Under this normalization scheme data with more positive contribution terms have a larger (negative) impact on the model predictor.
\begin{equation}\label{eq:econtributionnorm}
\phi^c_{i,j} = -\text{sgn}(e_i) * \phi_{i,j}
\end{equation}
Figure \ref{fig:iexample} illustrates the composition case of an individual instance, the total positive ($x_1,x_2$) and negative ($x_3, x_4, x_5$) effects which are contributing to the final residual of instance $x_3$. 
\\
\\
Influence typically measures the impact of an instance on some quantity of interest, such as the model parameters or goodness of fit statistics \cite{cook1979influential}. Contribution can be interpreted as the influence of an instance on the model-predicted outcome over every instance. In many cases, a large contribution value means that an instance has a large effect on the loss of the model (i.e. if the average residual is close to 0). Taking this aggregate over the data-set (the global contribution) yields  a similar quantity to the DataShapley value \cite{ghorbani2019data} which measures the `value' of an instance. This is because the total loss as measured by DataShapley is composed of the $n$ individual residual terms (i.e $L = \sum_i^N e_i^2$) which we individually measure here. On a local level, the contribution effects can be interpreted as the asymmetric pairwise interactions between instances with application to determine the effects of adding or removing instances.
\\
Our approach to computing $\phi_{i,j}$ only involves changing the definitions of the value function in previously well-known Shapley implementations and so inherits similar (the Shapley) properties and can be empirically verified. Most importantly is that the sum over compositions is equivalent to the residual value. There are several derived quantities from this definition that we may be interested in, for example how much an instance contributed to itself compared to others, the total residual of an instance (how many other instances are acting upon it), and other statistical quantities derived from the composition-contribution matrix. 
\section{Symmetric Applications and Results}

The symmetric case of Residual Decomposition occurs when the training set is the same as the testing set, in doing so we generally analyze the patterns that data exhibits within the entire sampled population, and can compare across populations and models (i.e. for data visualizations).

\subsection{Contribution-Composition (CC) Plots and Global Interpretation}

\begin{figure}[!ht]
    \centering
    \includegraphics[width=\textwidth/2]{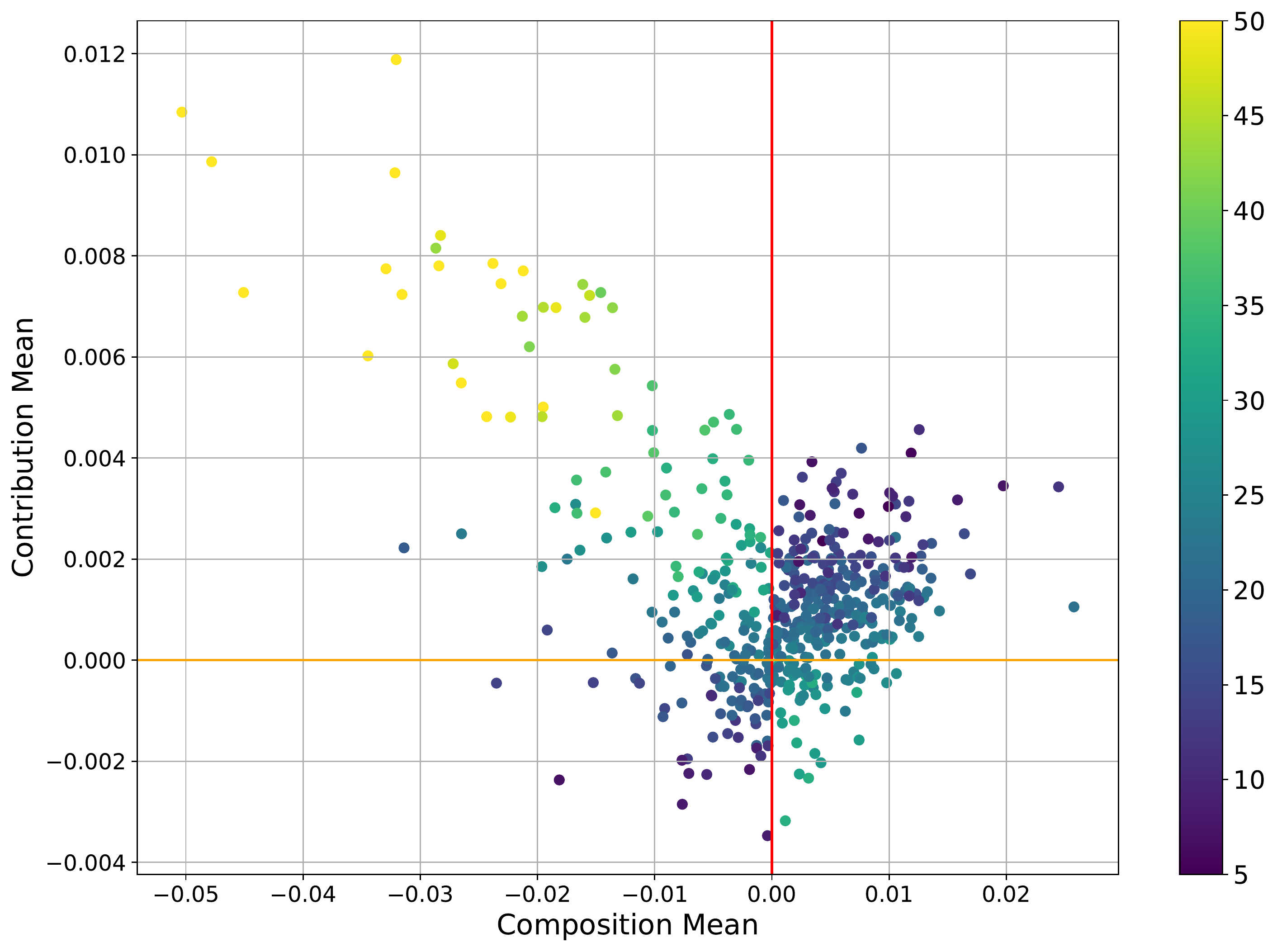}
    \caption{Composition vs Contribution plot for the Boston Housing data-set \cite{harrison1978hedonic} with Ridge Regression as the choice of model function and instances colored by median house value. Note: we use a corrected version of this dataset (removal of some features) due to recent ethical concerns \cite{scikit-learn}.}
    \label{fig:CCBoston}
\end{figure}

We begin by graphically visualizing the Shapley Values of residuals by plotting the summary mean of each axis of Composition $\phi_{i,j}$ and Contribution values $\phi_{i,j}^c$ on Figure \ref{fig:CCBoston} which we call a CC plot. This plot is generated by fitting some model to the data and plotting the properties of the data with respect to that particular model. The CC plot provides a global interpretation of the trends present in the data and allows us to observe the instances that deviate from these trends. It is immediately obvious that instances that deviate significantly from the group are candidates for outliers or require further investigation. We can see groups of `abnormal` instances, possibly outliers in figure \ref{fig:CCBoston}, the yellow instances with median house value approximately equal to $50$ have CC properties far from the rest of the data. It is well known in the Boston Housing data-set that some of the instances with a median value of $50$ are anomalous since there has been a truncating process carried out \cite{gilley1996harrison} and ignoring these instances results in a much more symmetric view of the data. Furthermore, these instances tend to have a high contribution value meaning that on average they tend to drive the average residual value upward. It is also the case that for regression-based models, instances that lie further away from others (i.e. high leverage points) tend to have a greater effect on the model \cite{cook1982residuals}. We also observe individual instances that lie far away from the main group despite having similar Y-values, it may be of interest to analyze these samples further. We provide additional CC plots on several machine learning data-sets in Appendix \ref{app:CCData} where the differences between the Ridge and Random Forest regressors are even more pronounced.
\begin{figure}[!t]
    \centering
    \includegraphics[width=22em]{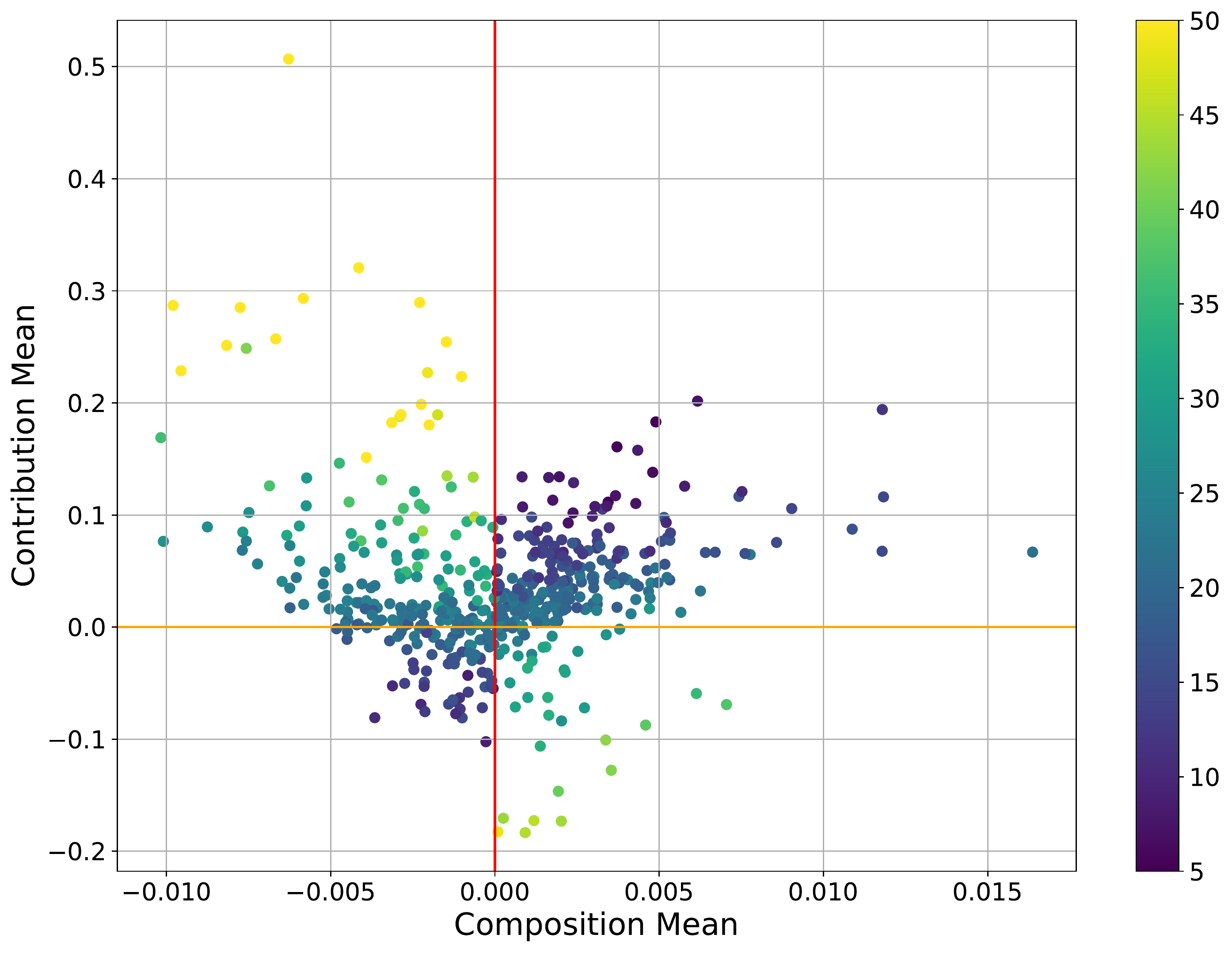}
    \caption{Composition vs Contribution plot for the Boston Housing data-set \cite{harrison1978hedonic} with a Random Forest Regressor (100 trees) as the choice of model function and instances colored by median house value. }
    \label{fig:CCBostonRF}
\end{figure}
\\
\\
This form of visualization of the data in the CC plot may serve as a general-purpose data visualization strategy that transforms any given data into the two dimensions of contribution and composition which have a consistent meaning across models and data. This contrasts with existing dimension reduction approaches to data visualization which produces subspaces that lack inherent meaning such as PCA or TSNE \cite{pearson1901liii, van2008visualizing}.  This is because the axes produced by those approaches are arbitrary based on the data properties themselves. Our method is only parameterized by the choice of model used to fit the Shapley values, it is therefore possible to compare different data-sets and models using our method to observe differences in the patterns of instance behaviors. We can also draw an analogy to the analysis of residuals, in statistics it is typically desirable for many classes of models to have uniformly distributed residuals. The CC plot conveys this information through the symmetry of each range of Y-values across the contribution mean axis. Additional desirable properties may be symmetric in the contribution of instances across a range of Y-values. Intepretation-wise, CC-plots can be used as an assessment of data and model quality, that is a desirable model tends to have low and uniform values for the residuals and can be expressed through the composition quantity axis. It is also desirable to have few outliers and instances which have significantly different effects to others within the data-set, it is therefore desirable to have biased distributions of contribution towards zero (i.e. Gaussian distribution) expressed through the contribution axis. Additionally, figures illustrating poor fit of models with the data can be seen in Appendix 
\\
\\
We can compare the CC plot in Figure \ref{fig:CCBoston} with the one present in Figure \ref{fig:CCBostonRF} trained using a significantly more powerful model and has lower composition values on average. We see that the Random Forest attempts to correct for the outlying yellow instances but is not able to completely do so since the majority of such instances remain with large contribution values. At the same time, the lower end of the Median values (purple instances) have become significantly more asymmetric. Together this suggests that neither are particularly appropriate models for the underlying data or that there is something inherently anomalous in the data (which may be revealed via the CC plot).
\begin{figure}[!th]
    \centering
    \includegraphics[width=0.5\textwidth]{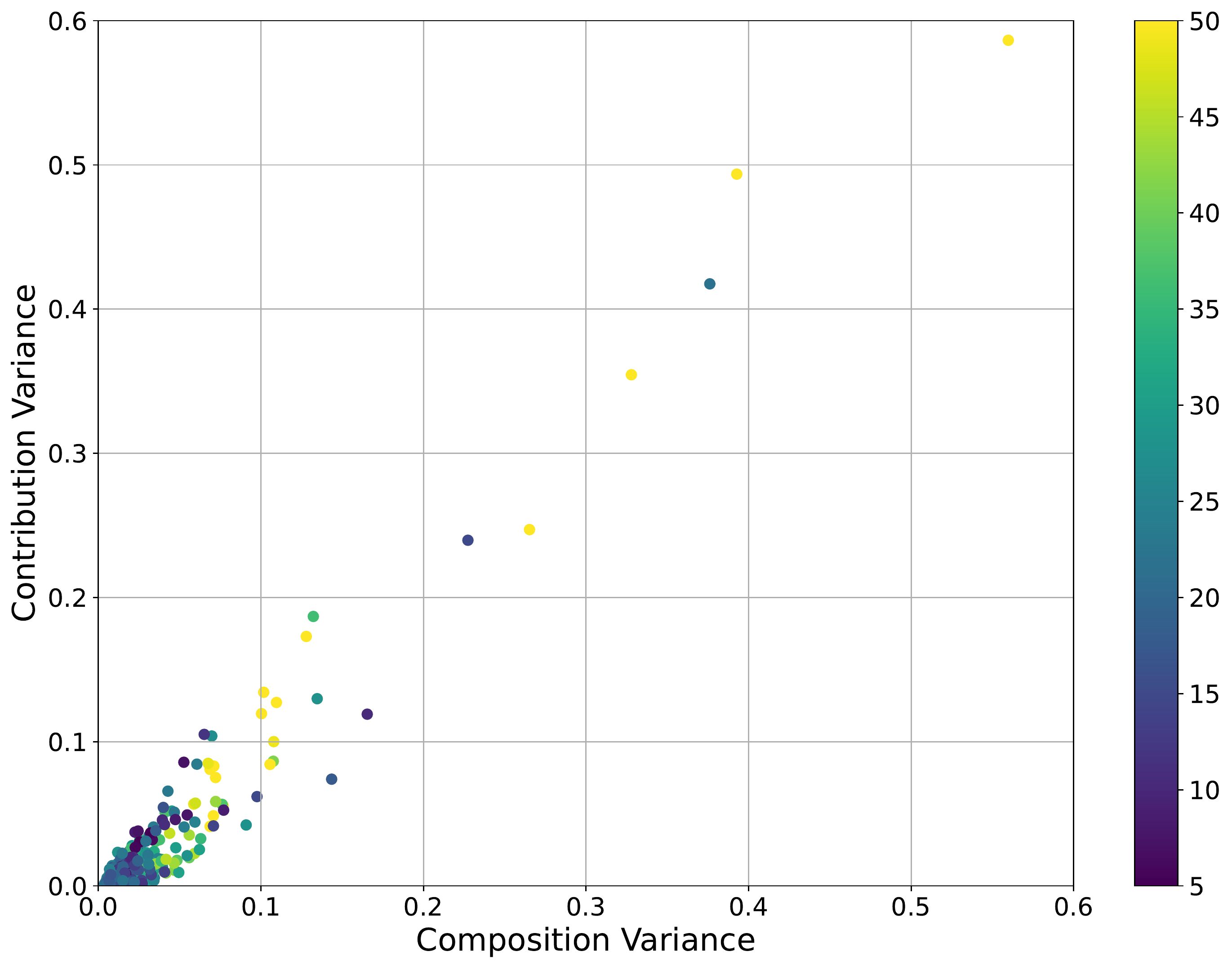}
    \caption{CC Plot (Variance) for the Boston Housing data with a Random Forest model.}
    \label{fig:CCBostonvar}
\end{figure}
Furthermore, additional separation of unique samples can be detected using higher order summaries or statistics on the $\bm{\phi}$ matrix. Figure \ref{fig:CCBostonvar} shows the CC-variance plot showing how the effects of individual instances differ across the sample dataset. It can be seen that while all the yellow instances in the CC-mean plot have similar effects, the particularly interesting points are separated out on this variance plot. Additionally, well-established techniques such as outlier detection or clustering can also be applied to separate out groups of instances. 

\subsection{Outliers, Force Plots, and Local Interpretation}

\begin{figure}[!t]
    \centering
    \includegraphics[width=\textwidth/2]{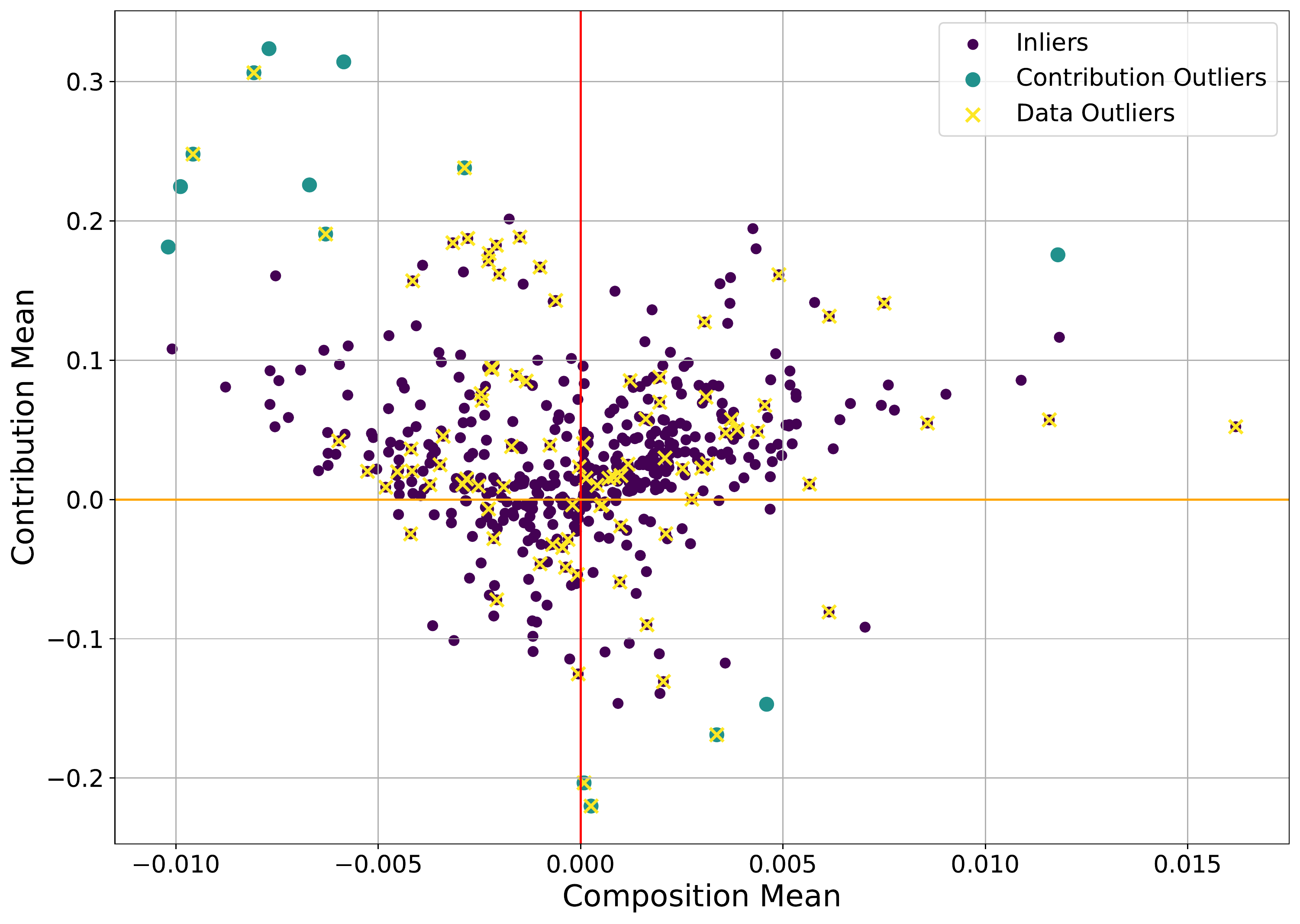}
    \caption{CC Plot based on Random Forest of outliers detected based on Isolation forest trained on contribution values, and raw data features.}
    \label{fig:bostonoutliers}
\end{figure}
In this section, we discuss how we can interpret Contribution and Composition on a local level for individual instances. While considering instances with interesting aggregated mean values is a good starting point for identifying instances of interest, individual instances may have interesting properties as well. For the majority of data-sets found in machine learning tasks, it is difficult to analyze every individual sample. Where such difficulties exist, well-established methods to determine deviating instances from the majority may be used such as techniques from outlier detection or Statistical analysis.
\\
\\
We begin with the detection of outliers in the context of contribution and composition. Our framework introduces a different approach to detecting outliers in the data by applying outlier detection methods to attributes that describe instance \textit{behaviors} (such as contribution and composition) instead of \textit{properties} (such as traditional metrics based on the data features). This is because data that deviate from each other are not necessarily outliers or problematic instances in the broader global context of the entire data-set and model. In many domains, we may not necessarily be interested in the outliers themselves, but rather in how they may affect given analyses. For example, does it make sense to remove exceptional individuals from an analysis if they provide some useful understanding of information? By analyzing contribution and composition in this way, previously unknown patterns which cannot be derived purely from features may be uncovered.
\\
\\
An example well-established outlier detection method is that of the Isolation Forest \cite{liu2008isolation} which determines how many `cuts' it takes to `isolate' an instance from the rest of the data. We apply an Isolation Forest to the contribution values and normal data seen in Figure \ref{fig:CCBostonRF} and produce the outliers seen in Figure \ref{fig:bostonoutliers}. While there are a significant number of detected outliers in the data based on their features, they do not have significant enough differences (according to the isolation forest) based on their behaviors to be considered outliers. It may also be appropriate to consider outliers based on data features, but only those which have had a significant effect on the model (i.e. high contribution values) since typically we are interested in only outliers that matter \cite{andrews1978finding}. Taken this way, our framework can be a useful tool in narrowing down the outliers that need to be treated. 
\begin{figure}[!t]
    \centering
    \includegraphics[width=\textwidth/2]{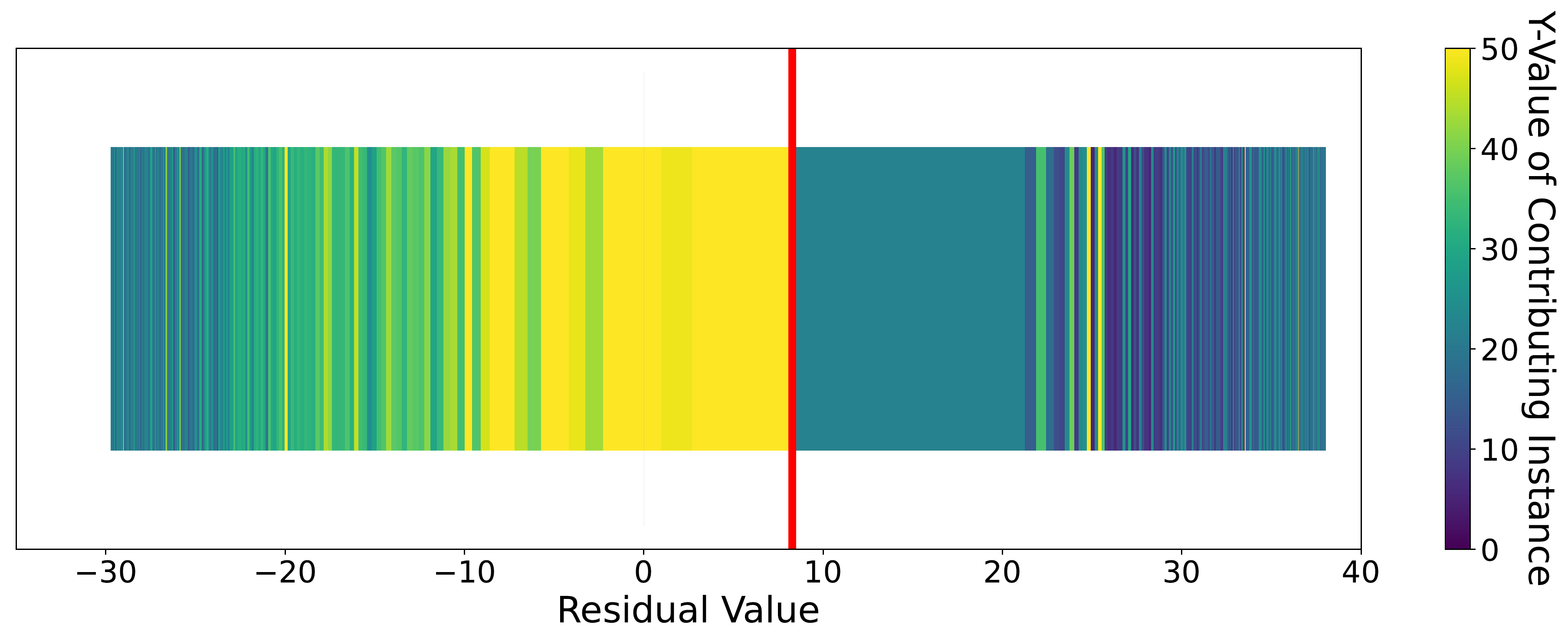}
    \includegraphics[width=\textwidth/2]{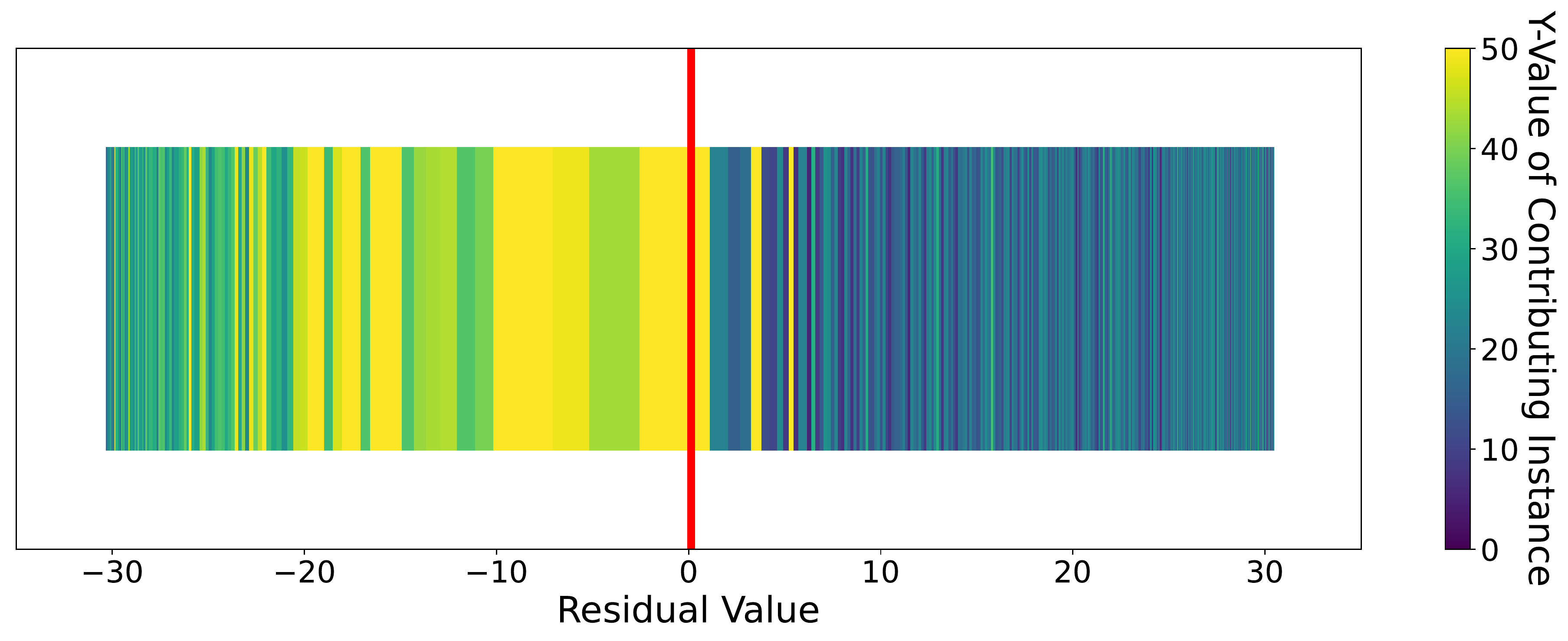}
    \caption{(Top) Force plot for a single instance with the largest residual value in the data-set (8.2). (Bottom) Force plot for a single instance with the smallest residual value in the data-set (0.001) based on Random Forest model.}
    \label{fig:ForceBoston}
\end{figure}
\\
\\
In order to individually analyze the composition of instances, we adapt force plots \cite{lundberg2017unified} which are a method of visualizing how much each feature contributes to the final predicted value of the model to our contribution of residuals case. Figure \ref{fig:ForceBoston} shows the force plot of the largest and smallest residual instances in the data. The direct interpretation of this plot is the same as that shown in Figure \ref{fig:iexample}. This force plot over instances conveys different information to the SHAP case since the total number of instances is so large and each individual contribution is significantly smaller. We rely on the color trends within the force plot to convey information, in this case, we can observe that the majority of the residual error can be attributed to the yellow instances with higher Y-values. If we wish to reduce the error of the first shown instance, then a large portion of the yellow (y=50) instances must be removed but this will also result in a decrease in the residual of the second instance. This sort of analysis may be useful when we want to identify why particular instances are mispredicted, along with quantifying the impacts of removing particular instances on others. 
\begin{figure}[!t]
    \centering
    \includegraphics[width=\textwidth/2]{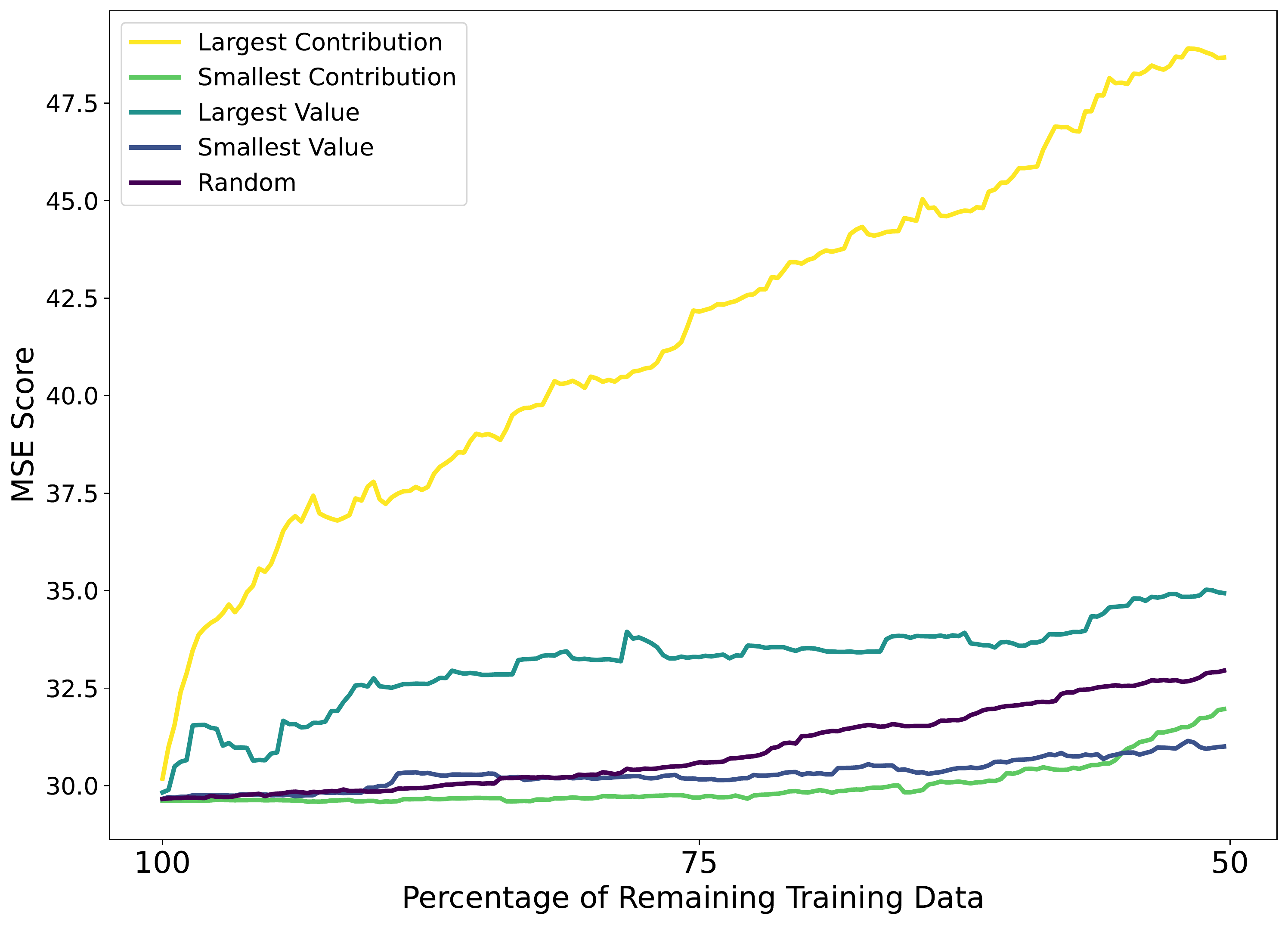}
    \includegraphics[width=\textwidth/2]{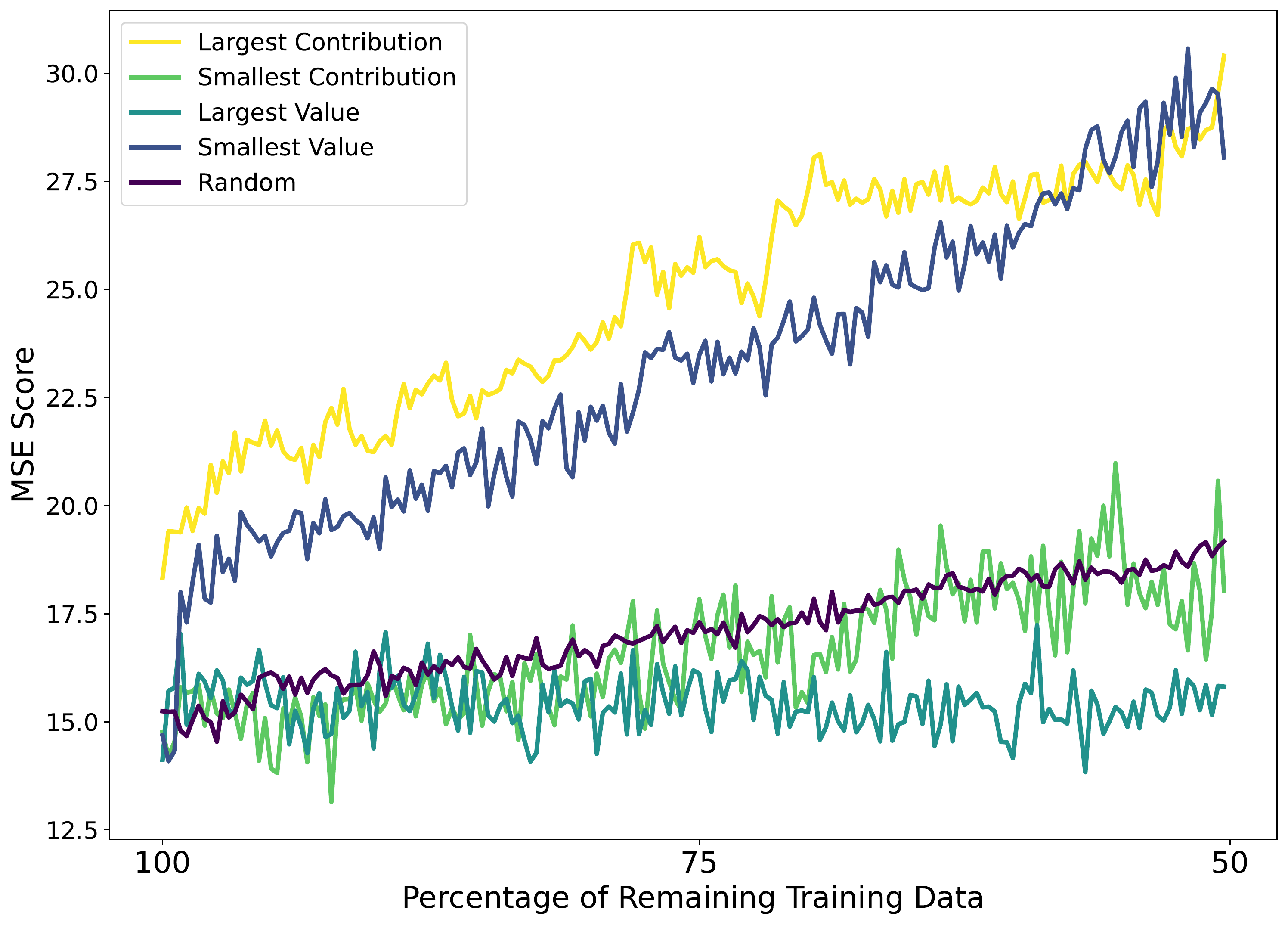}
    \caption{Mean Squared Error (MSE) of the testing set loss when removing instances based on Data Shapley Value and Residual Decomposition Contribution and trained using a Ridge Regression model (Top) and Random Forest (Bottom). Random selection was generated over repeated trials (n=10). Contribution produced by Residual Decomposition and Value produced by Data Shapley.}
    \label{fig:dshapvsrshap}
\end{figure}
\section{Asymmetric Applications and Results}

The residual decomposition method when applied with a training and testing set methodology produces a different interpretation to that of the symmetric case since the composition matrices are no longer square but rather $N_{\text{train}}$ by $N_\text{test}$. 
\\
\\
The total contribution of an instance in the training set to instances in the testing set is similar to the quantity measured by Data Shapley and can be used for the task of data valuation. This is because the loss function captured by Data Shapley values is precisely composed of the residuals of the model that is captured by contribution. The equitable properties of data valuation \cite{ghorbani2019data} may not be fully satisfied when using contribution since some aggregation has to be carried out to obtain a `value' for a particular instance. Despite this, we believe that the axiomatic underpinnings of Shapley Values are sufficient that any aggregation in this sense is intuitive and `similar' to the value produced by DataShapley. In the cases that an instance has a large Data Shapley value (meaning that it tends to increase the loss), we can generate a similar contribution value based on the functional form of the residual (i.e. the mean squared contributions of an instance). In Figure \ref{fig:dshapvsrshap} we compare the performance on the testing set if we remove instances from the training set based on the contribution assigned by Residual Decomposition and that of the Value assigned by Data Shapley. We select the value function for Data Shapley based on the Mean Squared Loss value, we take the squared mean of contribution to obtain the contribution equivalent of `value'. We see similar but different behaviors when removing the largest and smallest values with both methods.
\\
\\
Similar to the symmetric case, it is also possible to carry out analyses over contribution and composition such as cluster/outlier analyses and force plots. If we are interested in differences between predicted and mispredicted instances, we may be able to look at the force plot and see which regions are the causes of large portions of the residuals. Clustering and outlier detection techniques may also reveal individual or groups of anomalous data.

\begin{figure*}[!t]
    \centering
    \includegraphics[width=\textwidth]{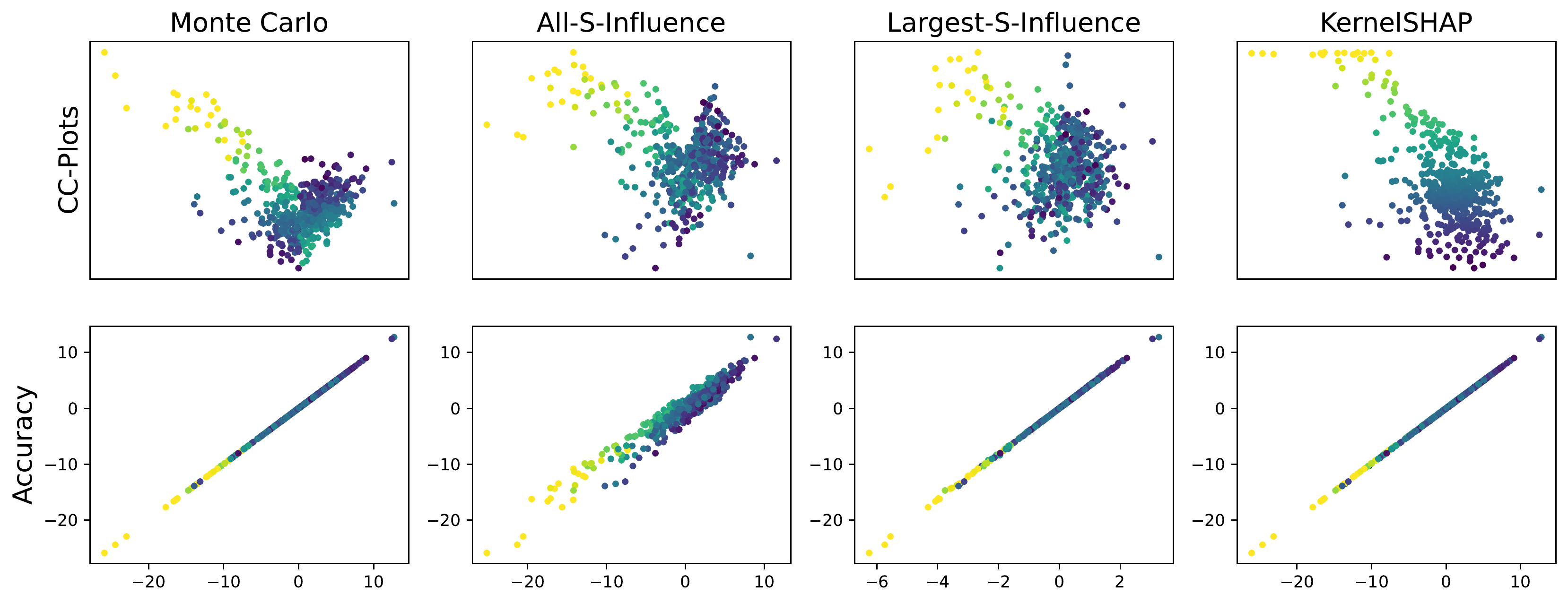}
    \caption{Comparison of CC plots and accuracy of various algorithms for computing Shapley Values of Ridge Regression model. Accuracy is measured as the actual residual of the model against the composition sum. }
    \label{fig:bostoncomparisons}
\end{figure*}

\section{Performance and Runtime Considerations}\label{sec.runtime.performance}

\begin{figure}
    \centering
    \includegraphics[width=\textwidth/2]{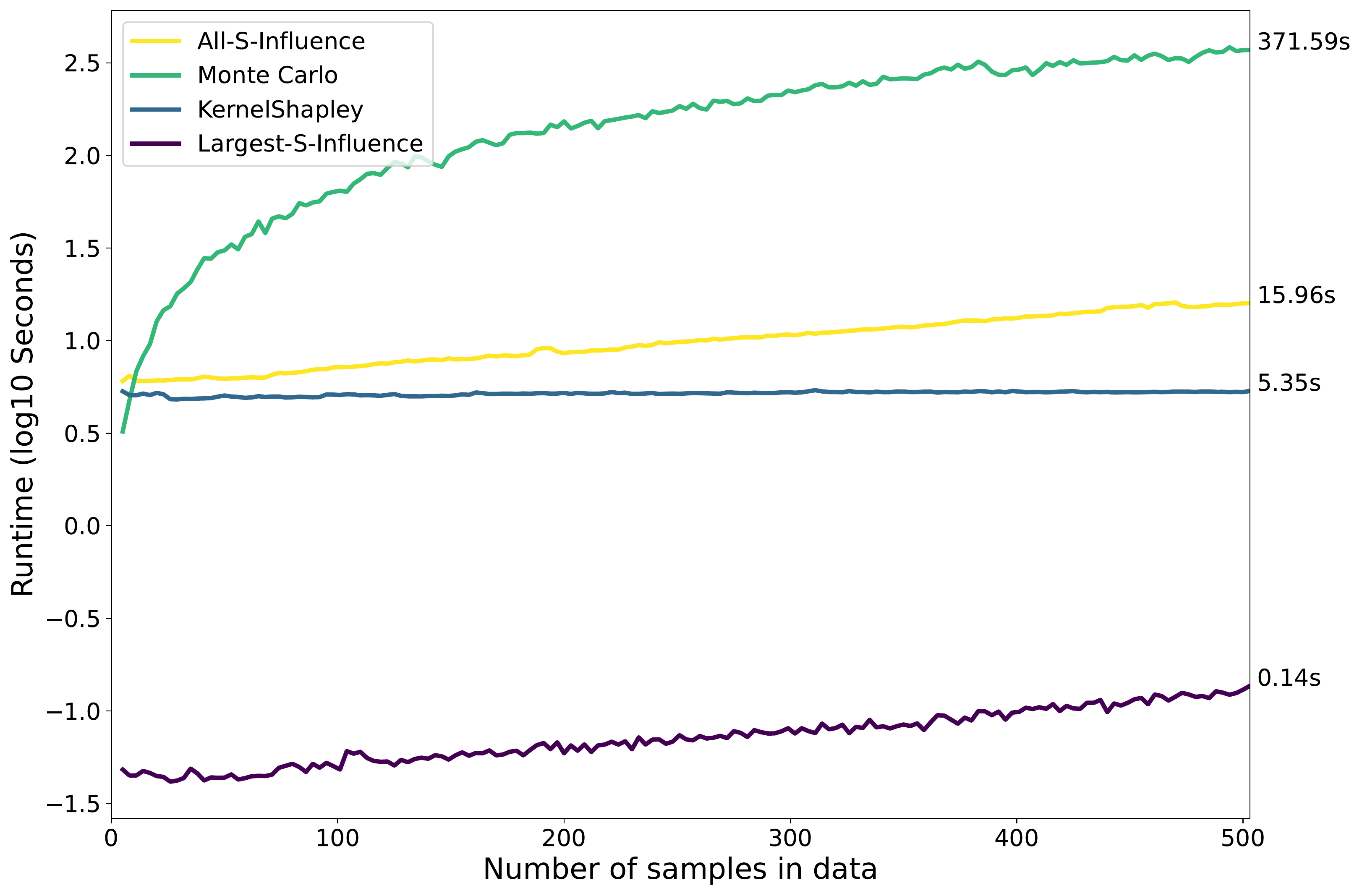}
    \caption{Runtimes for the four tested algorithms in $\log_{10}$ seconds and final runtimes evaluated with the entire dataset annotated on right.}
    \label{fig:bostonruntimecomparisons}
\end{figure}

In this section, we consider run-times and algorithmic challenges that need to be overcome. As the number of instances or the model complexity increases, the baseline permutation sampling algorithm quickly becomes infeasible in the run-time. For the intended purpose of this framework and working with small data-sets and simple models, Monte Carlo sampling typically suffices. The total run-time for the Monte Carlo permutation sampling with the Random Forest (n=100) in the symmetric case was 14 hours on an i7-10700 @ 2.9GHz, while the Ridge Regression took approximately 6 minutes illustrating the significance of model complexity on the run-time of Shapley based approaches over instances. It is notable that parallelization can see a significant speed-up up to the order of the number of iterations and can be carried out over multiple processors or machines \cite{ghorbani2019data} since the problem is embarrassingly parallel. However, small data-sets are typically not the case for the majority of machine learning tasks. While we mainly focused on applications for small data-sets and can have applications to many applied tasks such as those found in Health Informatics and Materials Sciences \cite{barnard2019nanoinformatics}, further significant speed-ups need still be achieved. 
\\
\\
For many classes of algorithms with an empirical risk minimizer of the form $\arg\min_\theta \frac{1}{n}\sum_{i=1}^n L((x_i, y_i), \theta)$ (such as Neural Networks), influence functions can be applied to significantly decrease the run-time to get an approximate solution. Jia et al. proposed using influence functions as a heuristic to improve the baseline Monte Carlo approaches\cite{doshi2017towards, jia2019towards}. This is because given some data $S_1$, then evaluating $U(S)$ gives $U^(S_2 \cup \{i\}) - U(S_2)$ for all $i \in S$ where $S_1 = S_2 \cup \{i\}$ which is the quantity of interest in computing the Shapley Value. Furthermore, \cite{jia2019towards} also proposed the largest-S algorithm which takes the influence over a single subset $I$ of the data. In a more general case for any model, KernelSHAP provides the provably optimal approach to sampling groups of instances/features to compute the Shapley Values. While additional speed-ups can be achieved by applying further constraints and assumptions, they are out of the scope of this paper and are left to future work. The alternative algorithms introduced by \cite{jia2019towards} also do not work well in our case since they involve solving an optimization problem in the number of Shapley values which is only $n$ in the Data Shapley case but $n^2$ in our Residual Decomposition case. Such approaches may yet still be faster than the baseline Monte Carlo algorithm, in our experience they still lag behind these influence-based approaches. For our influence experiments, we focus on the case of Ridge Regression since it has a closed-form second derivative which can be efficiently computed for the case of influence functions. The CC plots and accuracy of the four algorithms and the average time $t$ to produce the final outputs are shown in Figure \ref{fig:bostoncomparisons}. Accuracy is measured by the difference between the residuals of the model and the sum of the composition value produced by the Residual Decomposition. The total runtimes for each model of varying problem sizes can be seen in Figure \ref{fig:bostonruntimecomparisons} and it can be seen that the influence-based approaches scale significantly better than the baseline Monte Carlo approach.
\\
\\
The drawbacks of the influence-based approaches are in the accuracy of the approximation, while influence provides a very good estimate of the error when removing one training point, it is not completely precise as seen in Figure \ref{fig:bostoncomparisons}. In the case of the all-S-algorithm, these errors can compound over the various considered subsets. Several works have also critiqued the accuracy of influence functions (i.e. in the context of approximating leave-one-out error \cite{bae2022if}), thus further experiments are required in the context of residual decomposition to quantify exactly how much they differ and what are acceptable trade-offs of accuracy versus run-time. This is because explanations themselves are seldom fully correct and are typically approximate solutions to the true model. The largest-S-influence algorithm produces a CC plot close to the all-S-influence in less time, however only considering a single subset means that it is impossible to satisfy the additivity principle resulting in poor accuracy. Influence functions have recently seen significant success in contemporary machine learning and deep learning tasks \cite{koh2017understanding} since the second derivative is straightforward to compute within frameworks such as PyTorch or TensorFlow. However, despite the speed-ups from influence models, the general task of computing Shapley Values over instances remains difficult as it still requires a baseline number of models trained. While KernelSHAP works well in many feature importance explanation methods, as the number of permutations becomes large, KernelSHAP will only sample from $k=1$ and $k=N-1$ based on the Shapley Kernel weightings. These results and scaling to significantly larger data-sets indicate that significant algorithmic improvements may still be made in the context of the instance space explanations for faster and more accurate approximations to the true Shapley Value. Despite the generally poor approximations produced by these algorithms, they are still useful, notably the all-S-influence algorithm since significant separation of the effects of various instances can still be achieved for the purposes of instance analysis. It also should be noted that model-specific approaches can also produce significant speedups such as that of the TreeExplainer provided by the SHAP package \cite{lundberg2017unified}.

\section{Discussion}

We proposed Residual Decomposition as an alternative method for the task of data-set and instance analysis which can also be used for the task of data valuation in the asymmetric case. This framework which is based on the well-studied Shapley Values satisfies key desirable properties in the context of explaining how instances interact and ties into the literature regarding Data Valuation, Shapley Values, and Instance Analysis. Contribution and composition can be valuable tools to add to a toolkit in analyzing or comparing data-sets, making the correct model choices, and aiding in data visualization. Much like in the case of Data Valuation, context is important since the contribution and composition values lie within the context given by the data, model, and valuation methodology \cite{ghorbani2019data}. An instance with low contribution within one data-set may become important with the addition or removal of data, or a different choice of model and parameters.
\\
\\
This paper focuses on the case of Regression, in the future we plan to extend our work to the Classification case with an appropriate choice of the value function. The challenge is that there is typically little notion of distances and residuals in the context of classification meaning that some novel choice of value must be selected. The development of more performant algorithms must also be considered since scaling to larger datasets and more complex models is still a significant challenge.
\section*{Acknowledgements}

This research/project was undertaken with the assistance of resources and services from the National Computational Infrastructure (NCI), which is supported by the Australian Government, along with support by an Australian Government Research Training Program (RTP) Scholarship.

\section{Code}

Code for this project can be found at \url{github.com/uilymmot/residual-decomposition}.

\small{
\bibliographystyle{icml2023}
\bibliography{ref_final}
}

\clearpage

\onecolumn 
\appendix

\section{Additional CC-Plots on Different Data-sets}\label{app:CCData}

\subsection{Additional CC Data for Ridge Regression}

\begin{figure}[H]
    \centering
    \subfloat[\centering CC plot for the Boston Housing data-set \cite{harrison1978hedonic}.]{
    \includegraphics[width=0.22\textwidth]{Figures/CCBoston.pdf}
    }
    \subfloat[\centering CC plot for the Aquatic Toxicity data-set\cite{cassotti2014prediction}.]{
    \includegraphics[width=0.22\textwidth]{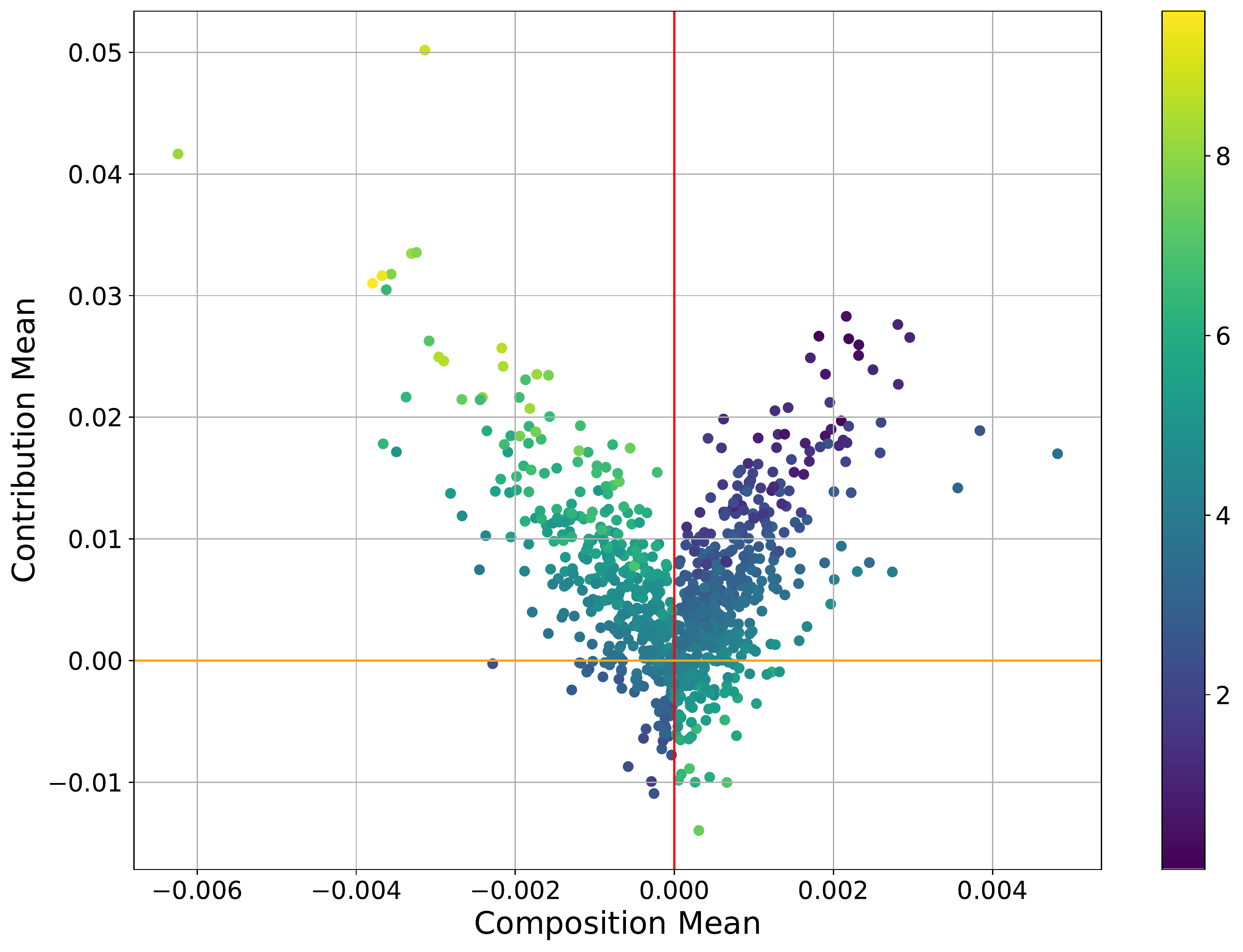}
    }
    \subfloat[\centering CC plot for the Compressive Concrete data-set \cite{yeh2007modeling}]{
    \includegraphics[width=0.22\textwidth]{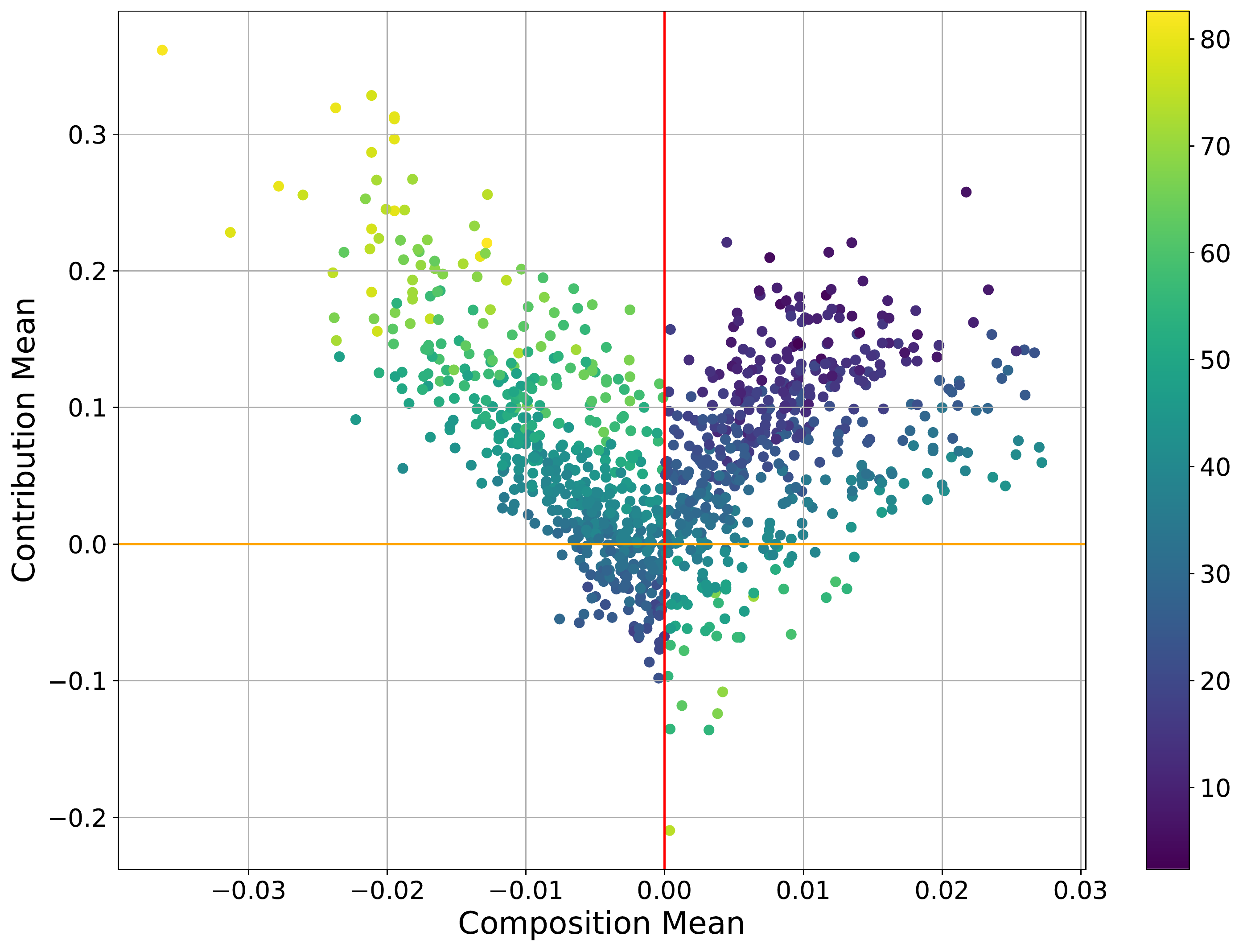}
    }
    \subfloat[\centering CC plot for the Graphene Oxide Bulk data-set (PCA components=10) \cite{go-data}]{
    \includegraphics[width=0.22\textwidth]{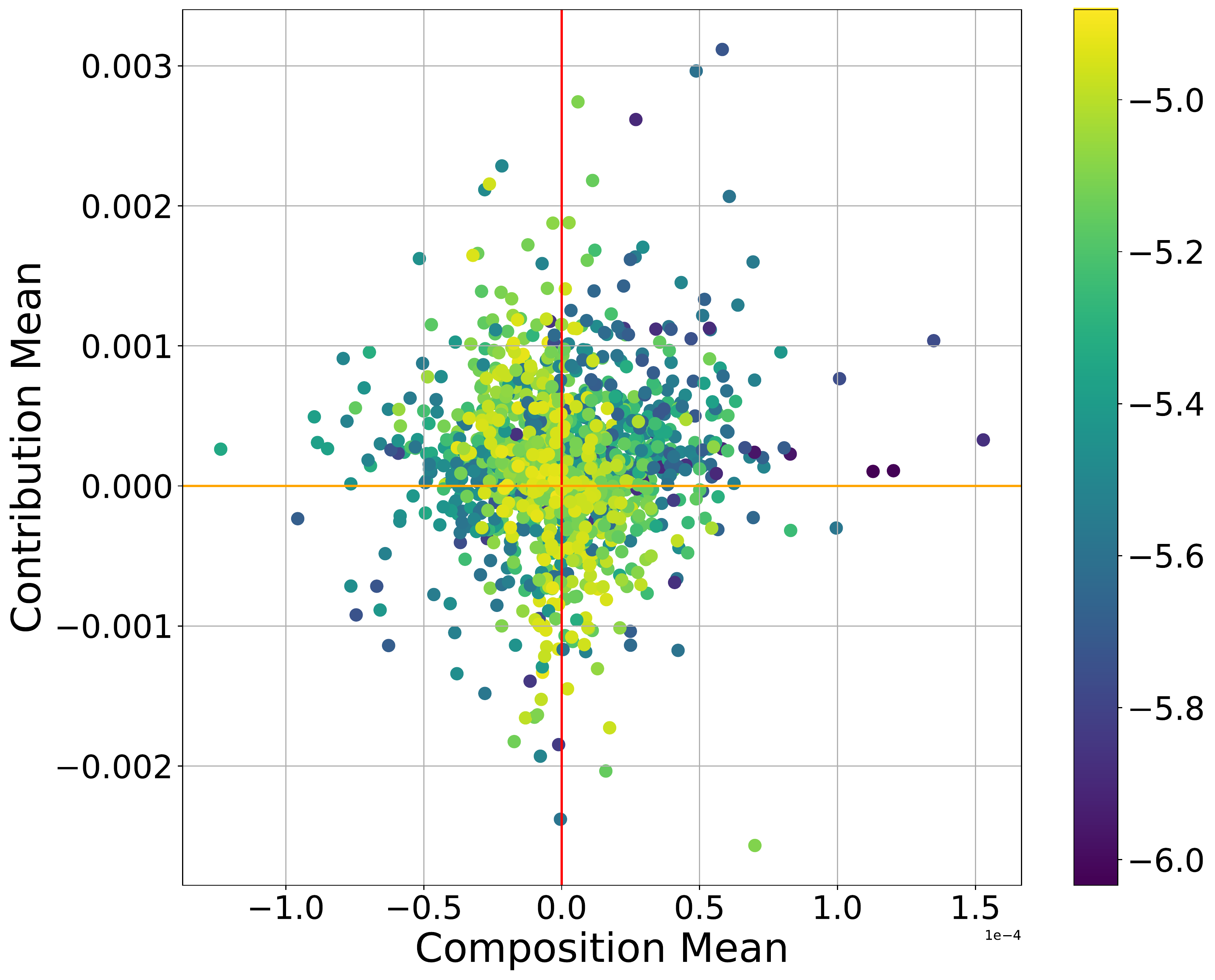}
    }
    \caption{CC Plots for various datasets using Ridge Regression model.}
    \label{fig:CCRidgeSUP_A}
\end{figure}

\subsection{Additional CC Data for Random Forest Regressor}

\begin{figure}[H]
    \centering
    \subfloat[\centering CC plot for the Boston Housing data-set \cite{harrison1978hedonic}.]{
    \includegraphics[width=0.22\textwidth]{Figures/CCBostonRF.pdf}
    }
    \subfloat[\centering CC plot for the Aquatic Toxicity data-set\cite{cassotti2014prediction}.]{
    \includegraphics[width=0.22\textwidth]{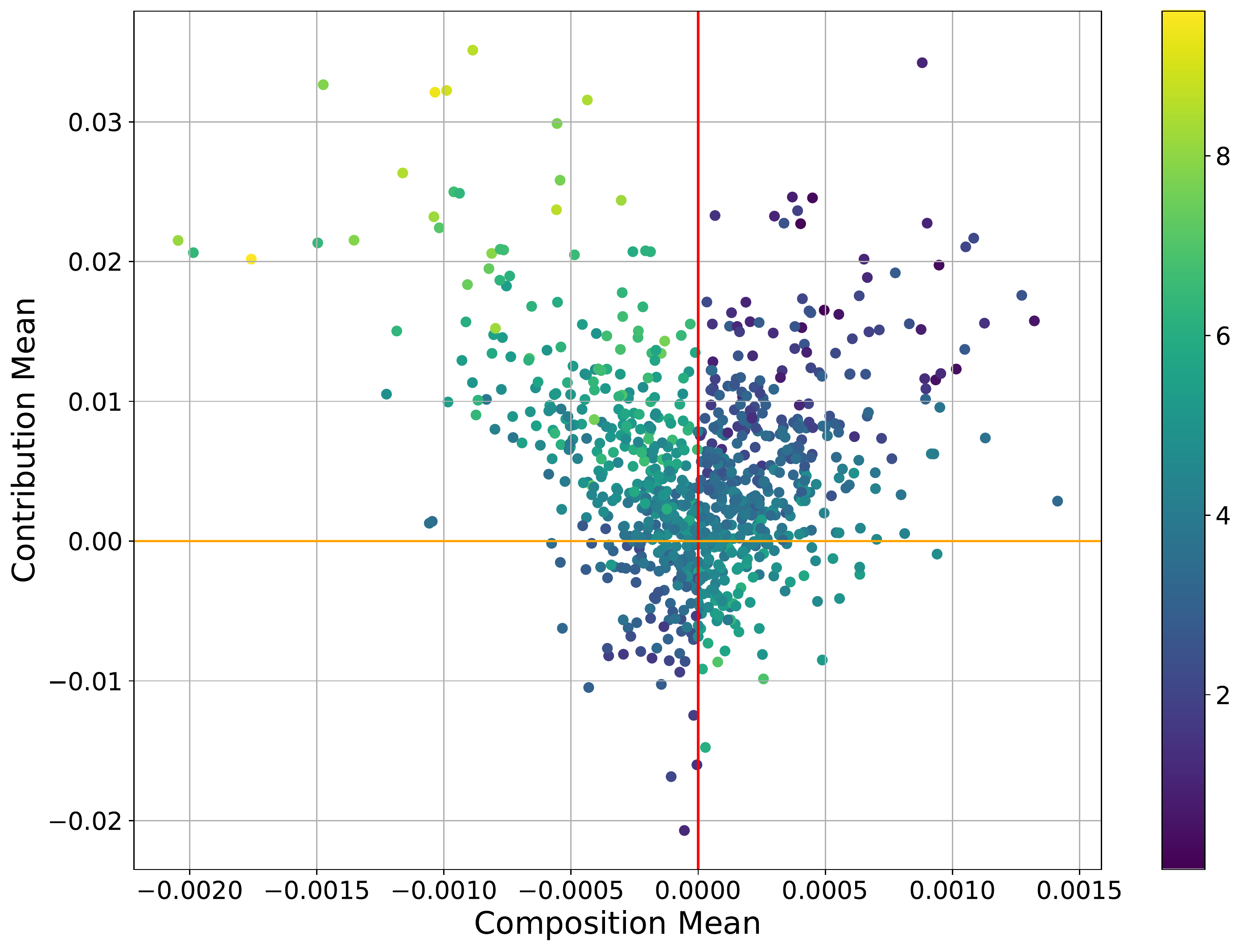}
    }
    \subfloat[\centering CC plot for the Compressive Concrete data-set \cite{yeh2007modeling}]{
    \includegraphics[width=0.22\textwidth]{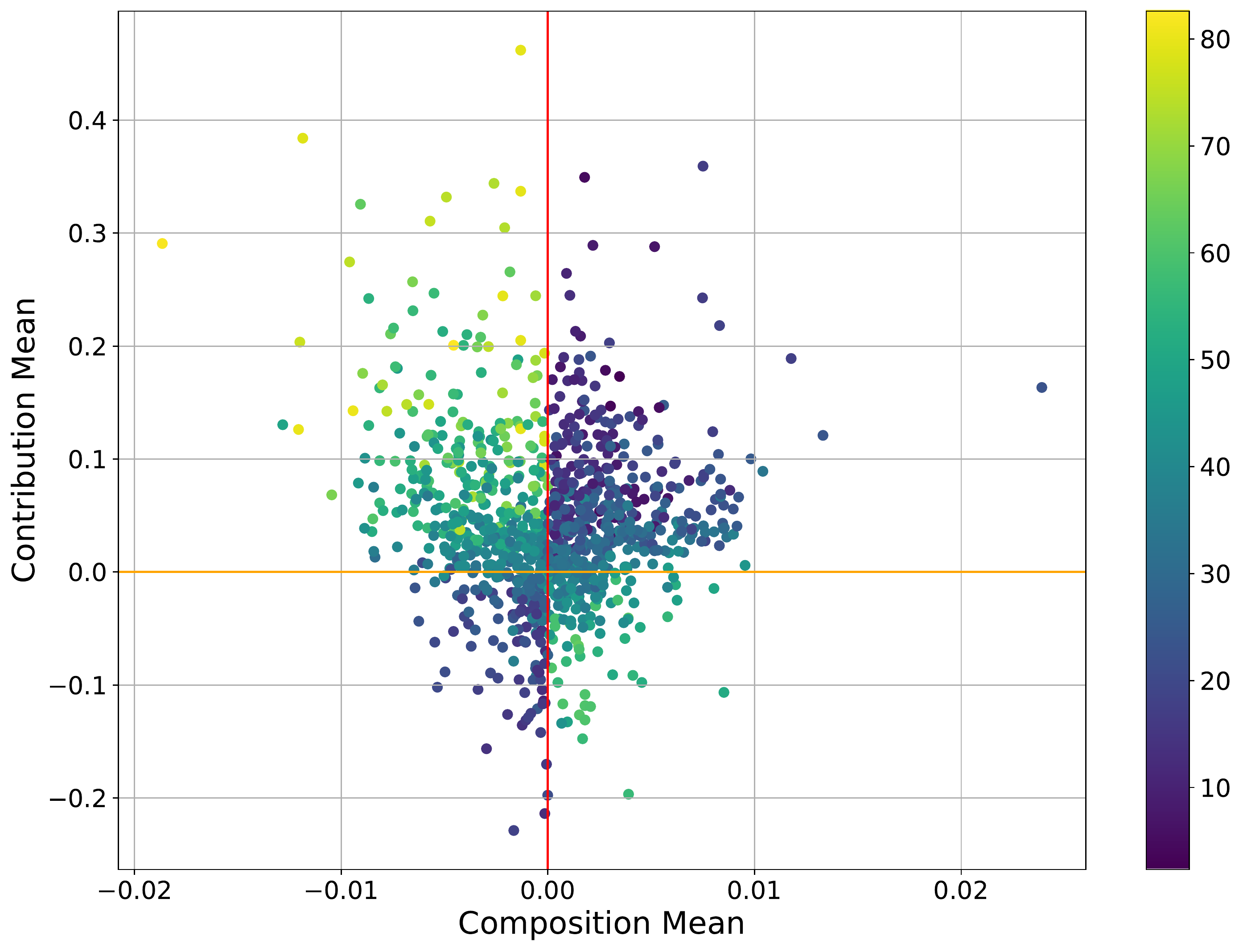}
    }
    \subfloat[\centering CC plot for the Graphene Oxide Bulk data-set (PCA components=10) \cite{go-data}]{
    \includegraphics[width=0.22\textwidth]{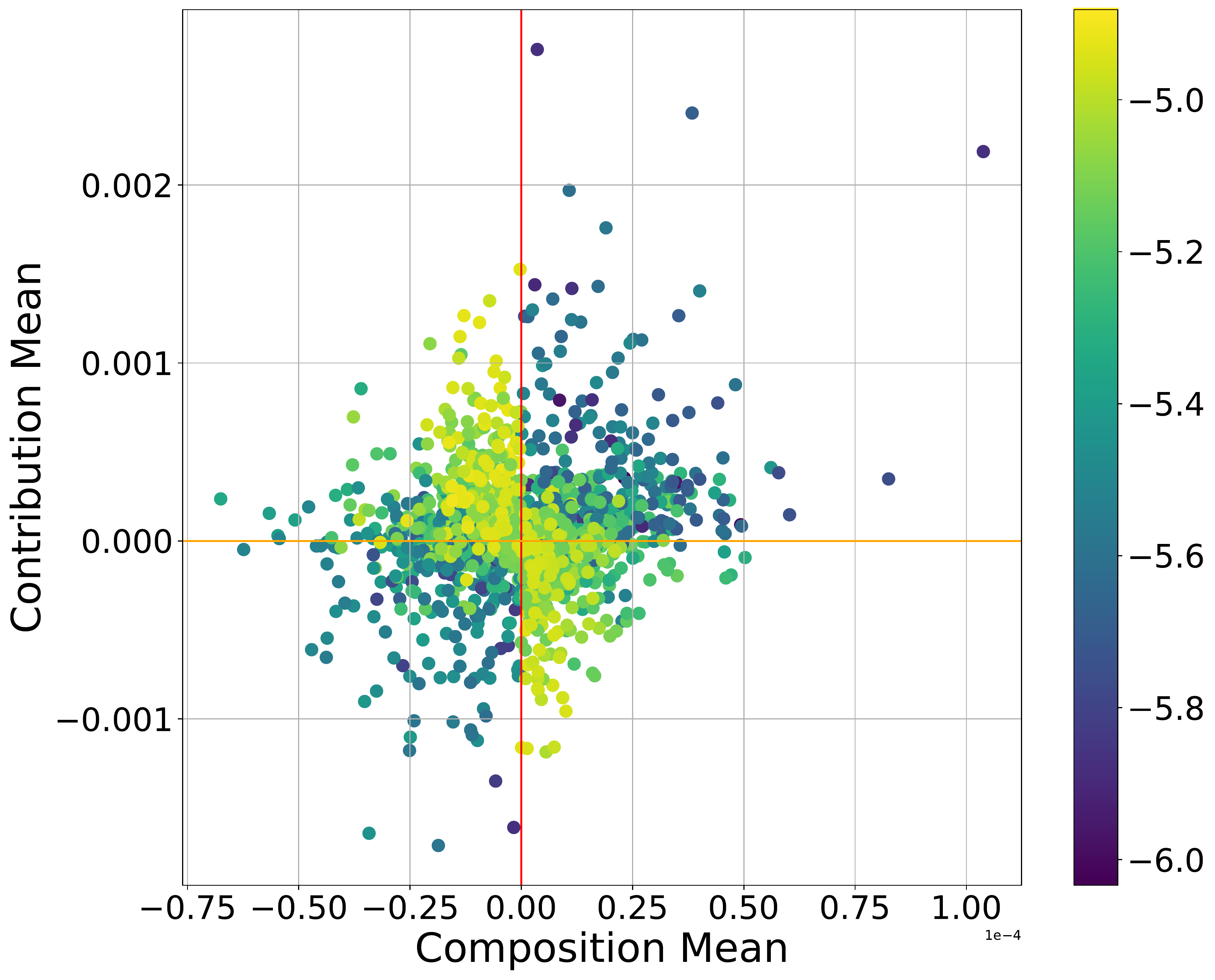}
    }
    \caption{CC Plots for various datasets using Random Forest model. We see significant regularisation effects compared to the above CC plots generated by the Ridge Regression models. Graphene Oxide data is highly regularised because of the PCA transformation as it reduces variances along the given principal axes. }
    \label{fig:CCRFSUP_B}
\end{figure}

\subsection{Additional Dataset Examples}

\begin{figure}[H]
    \centering
    \includegraphics[width=\textwidth]{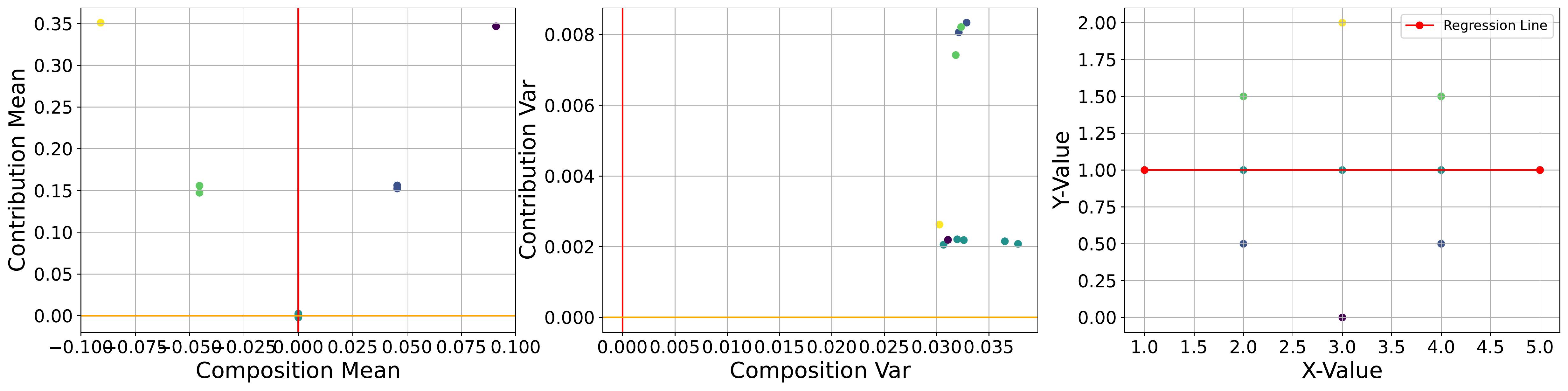}
    \caption{CC Mean and Variance plots of example 2 dimensional dataset (plotted on the rightmost plot). We see that the effects of the instances above and below the trend line are equal and opposite. It's also the case that instances further away tend to have higher impacts on the model. }
    \label{fig:toyA}
\end{figure}

\begin{figure}[H]
    \centering
    \includegraphics[width=\textwidth]{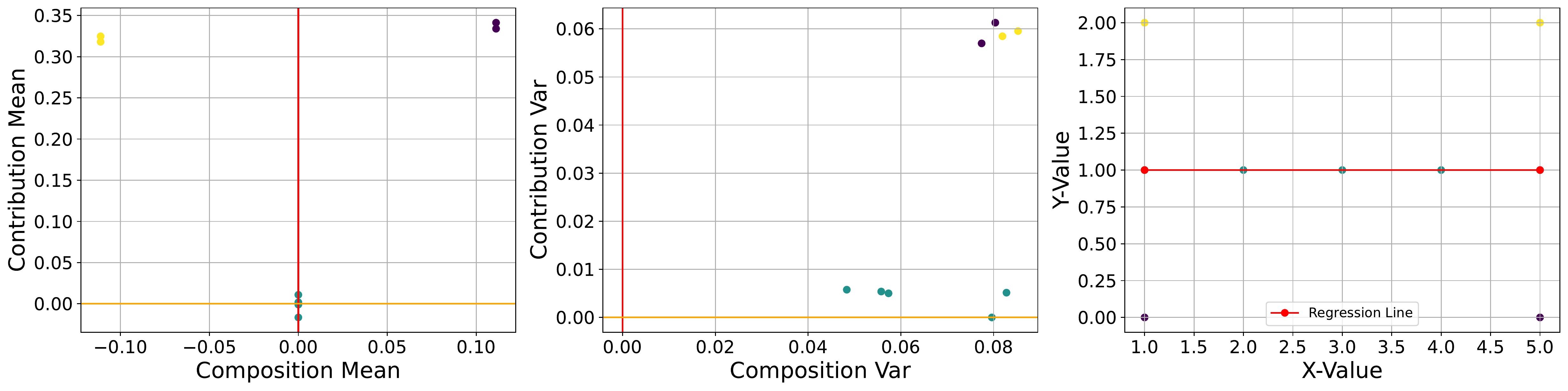}
    \caption{CC Mean and Variance plots of example 2 dimensional dataset (plotted on the rightmost plot). }
    \label{fig:toyB}
\end{figure}

\begin{figure}[H]
    \centering
    \includegraphics[width=\textwidth]{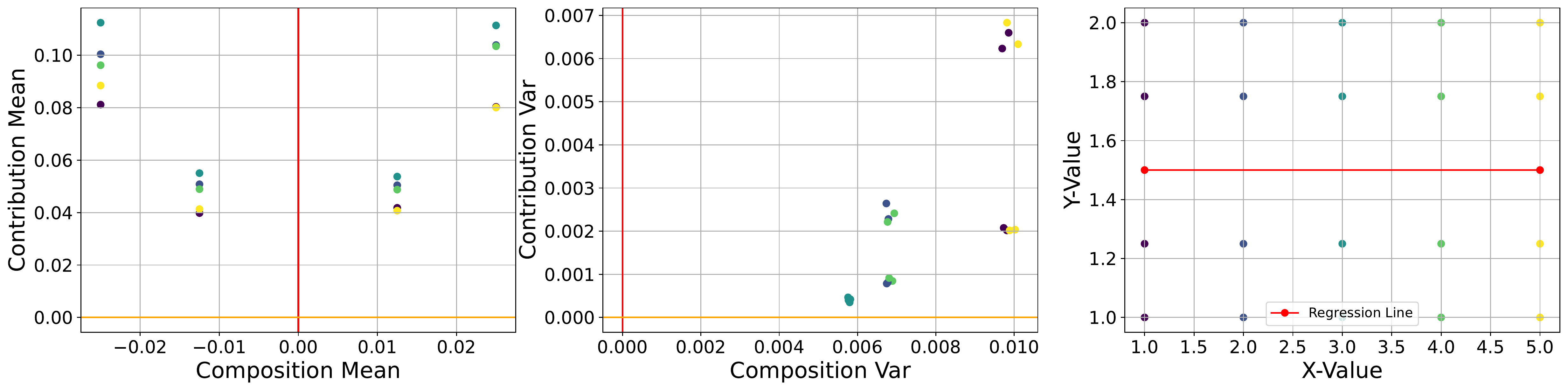}
    \caption{CC Mean and Variance plots of example 2 dimensional dataset (plotted on the rightmost plot). It can be seen that numerical and convergence errors are quite high for these small datasets since the likelihood of choosing the same or similar subsets through different permutations is high. }
    \label{fig:toyC}
\end{figure}

%
%
%

\end{document}